\documentclass[10pt,twocolumn,letterpaper]{article}

\usepackage[pagenumbers]{cvpr} 

\usepackage{graphicx}
\usepackage{amsmath}
\usepackage{amssymb}
\usepackage{booktabs}
\makeatletter
\@namedef{ver@everyshi.sty}{}
\makeatother
\usepackage{pgfplots}

\usepackage[pagebackref,breaklinks,colorlinks]{hyperref}

\usepackage[capitalize]{cleveref}
\crefname{section}{Sec.}{Secs.}
\Crefname{section}{Section}{Sections}
\Crefname{table}{Table}{Tables}
\crefname{table}{Tab.}{Tabs.}

\def\cvprPaperID{*****} 
\def\confName{CVPR}
\def\confYear{2023}

\begin{document}

\title{The 1st-place Solution for ECCV 2022 Multiple People Tracking in Group Dance Challenge}

\author{
  Yuang Zhang\textsuperscript{1,2},
  Tiancai Wang\textsuperscript{1},
  Weiyao Lin\textsuperscript{2},
  Xiangyu Zhang\textsuperscript{1}\\
  \textsuperscript{1}MEGVII Technology\\
  \textsuperscript{2}Shanghai Jiao Tong University\\
}
\maketitle

\begin{abstract}
  We present our 1st place solution to the Group Dance Multiple People Tracking Challenge.
  Based on MOTR: End-to-End Multiple-Object Tracking with Transformer~\cite{zeng2021motr}, we explore: 1) detect queries as anchors, 2) tracking as query denoising, 3) joint training on pseudo video clips generated from CrowdHuman dataset~\cite{shao2018crowdhuman}, and 4) using the YOLOX~\cite{yolox2021} detection proposals for the anchor initialization of detect queries. Our method achieves 73.4\% HOTA on the DanceTrack test set, surpassing the second-place solution by +6.8\% HOTA.
\end{abstract}

\section{Introduction}
The DanceTrack \cite{peize2021dance} dataset with uniform appearance is proposed to re-emphasize motion-based tracking.
Predominant approaches \cite{zhang2021bytetrack,cao2022observation} for MOT Challenge \cite{leal2015motchallenge, milan2016mot16} are mainly benefited from the tracking-by-detection pipeline with strong object detectors, such as YOLOX \cite{yolox2021}.
It partially results from the crowd scenarios with simple motion in the dataset and MOTA metric \cite{bernardin2008evaluating} bias towards detection performance.
As a result, multi-object trackers directly tuned on MOT Challenge fail to well model the motions in complex group dance scenarios.

In contrast, MOTR \cite{zeng2021motr} models tracked instances by ``track query'' and extends DETR \cite{carion2020detr,zhu2020deformdetr} for learnable multi-object tracking by iterative prediction. As indicated in the Table 3 of DanceTrack Paper \cite{peize2021dance}, MOTR achieved the best performance on AssA metric \cite{hota2021}, which shows promising potential to associate people in a complex group dance. However, its detection performance is inferior to those YOLOX-based trackers, like OC-SORT \cite{cao2022observation}.

\begin{figure}
  \centering
  \includegraphics[width=\linewidth]{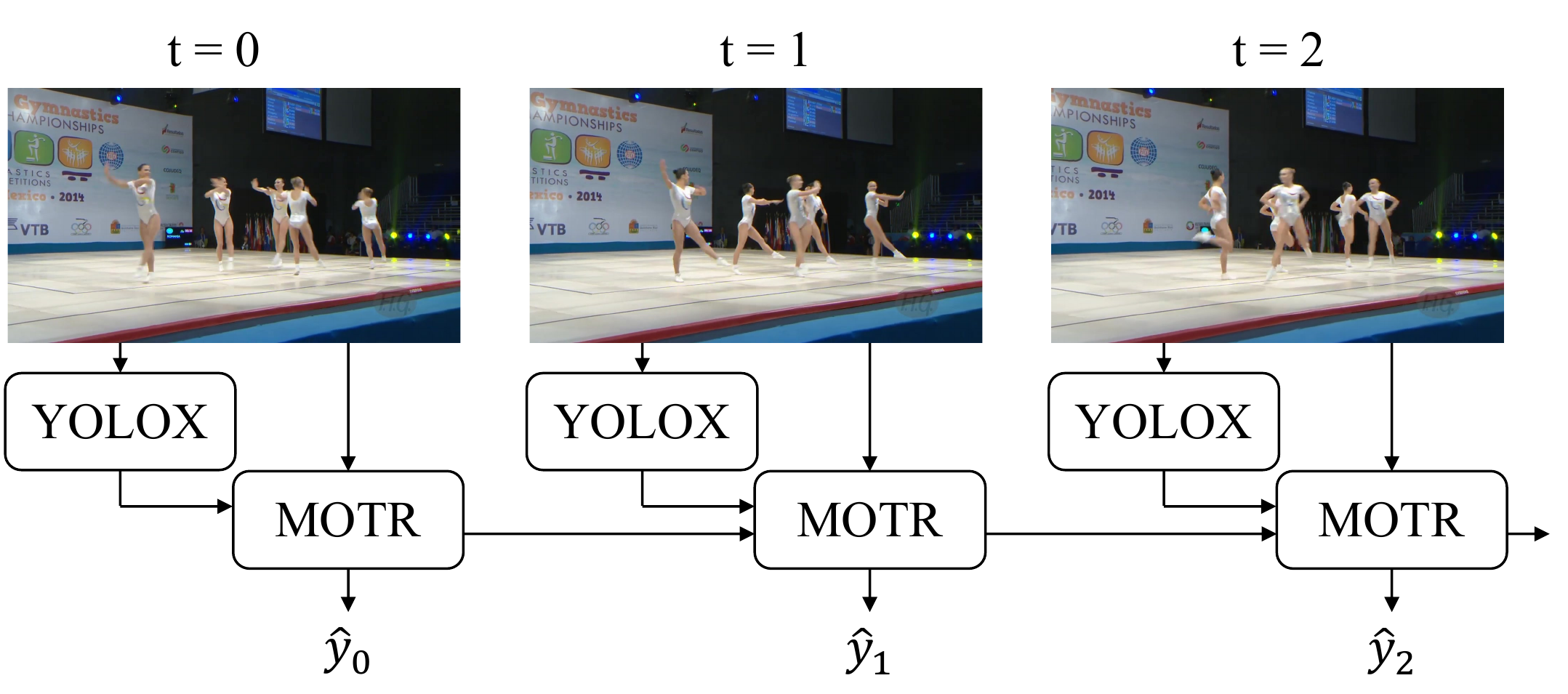}
  \caption{Overview of our solution. It combines state-of-the-art detector YOLOX \cite{yolox2021} and online tracker MOTR \cite{zeng2021motr}. YOLOX is employed to generate high-quality object proposals for each frame. The proposals are then used to provide a good initialization for detect queries of MOTR. It means detection process is decoupled from MOTR, which only performs the tracking association.}
  \label{fig:arch}
\end{figure}

Our goal is to achieve YOLOX-level detection performance and further extend the association capability of MOTR. We use YOLOX \cite{yolox2021} for proposal generation and the proposal boxes are used for the detect query initialization for MOTR \cite{zeng2021motr}. With the simple framework illustrated in \autoref{fig:arch}, our method outperforms the second-place solution nearly 10\% on AssA metric and achieves the second-highest DetA performance (only 0.02\% behind the solution with the highest DetA) in the group dance challenge.

\section{Method}
\label{sec:method}
As shown in \autoref{fig:arch}, the overall framework is based on MOTR \cite{zeng2021motr}. A pretrained YOLOX \cite{yolox2021} detector is used to generate high-quality proposals for the temporal association in MOTR. It combines their advantages on detection and association performance.

\subsection{Detect Queries as Anchors}
Multi-object tracking is a relatively local task since the moving distance of objects is usually small within short time interval. So it will be better to provide more local information for the detect queries employed in MOTR. Following Anchor DETR \cite{wang2021anchor} and DAB-DETR \cite{liu2022dab}, we replace the learnable positional embedding (PE) of detect query in MOTR with the sine-cosine PE of dense anchors, producing an anchor-based MOTR tracker.

\subsection{Tracking as Query Denoising}
DN-DETR \cite{li2022dn} by Li \etal introduces query denoising (QD) as a way of stabilizing bipartite matching in DETR \cite{carion2020detr}, simultaneously accelerating convergence and improving object detection performance. Li \etal feed the jittered ground-truth bounding boxes into DETR decoder and the decoder is trained to recover the original bounding boxes.

We observe that the query denoising task can be inherently similar to the task of tracking instances across frames: one bounding box prediction of previous frame can be seen as the current frame one with noise. Based on the observation, we adjust the query denoising in \cite{li2022dn} so that the scale of random noise added to the ground-truth boxes is comparable to the scale of object movement across frames for multi-object tracking.

Specifically, we choose the noise scale hyperparameters $\lambda_1=0.1, \lambda_2=0.05$ as opposed to $\lambda_1=\lambda_2=0.4$ in DN-DETR \cite{li2022dn}. For each ground-truth bounding box $(x, y, w, h)$, the center is shifted by a random noise $(\Delta x, \Delta y)$ where $\Delta x\sim \mathrm{Uniform}(-0.05 w, 0.05 w)$ and $\Delta y\sim \mathrm{Uniform}(-0.05 h, 0.05 h)$. The jittered width and height are uniformly sampled from $(0.95w, 1.05w)\times (0.95h, 1.05w)$. Note that the noise we choose has a much smaller scale compared to DN-DETR, and is more suitable for crowd scenarios and more consistent with the natural bounding box offsets across frames.

\subsection{Joint Training on CrowdHuman}
To enhance the detection performance, we use the images from CrowdHuman \cite{shao2018crowdhuman} dataset to generate pseudo videos for the joint training of MOTR tracker. Similar to the joint training of MOT17 and CrowdHuman in MOTR~\cite{zeng2021motr}, we generate pseudo video clips for CrowdHuman and perform the joint training with DanceTrack. The length of pseudo video clip is set to 5 for simplicity.

\subsection{Using YOLOX Proposals}
To further improve the detection performance, a pretrained YOLOX \cite{yolox2021} detector is used to generate the bounding box proposals. To combine it with MOTR in a elegant manner and keep the end-to-end feature of MOTR, we introduce a very simple pipline. The proposals generated by YOLOX are served as the anchor initialization of queries for the modified anchor-based MOTR tracker. The high-quality proposal provides a good initialization for object localization. The transformer decoder of MOTR predicts the relative offsets compared to the initial localization. With the employment of YOLOX, the detection and association processes of MOTR are fully decoupled and MOTR is only used to perform end-to-end association.

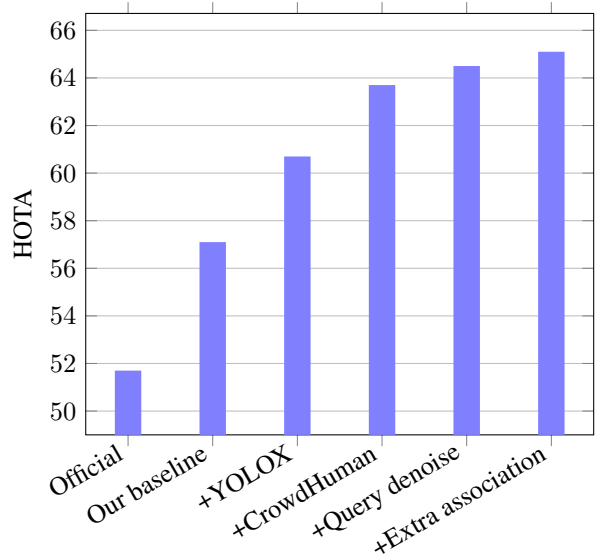
\begin{figure}
  \begin{tikzpicture}
    \begin{axis}[
        width=\linewidth,
        ybar,
        ylabel=HOTA,
        ylabel near ticks,
        ymajorgrids=true,
        ytick={50, 52, 54, 56, 58, 60, 62, 64, 66},
        ymin=49,
        xtick={0,...,5},
        xticklabels={Official, Our baseline, +YOLOX, +CrowdHuman, +Query denoise, +Extra association},
        x tick label style={rotate=30, anchor=east},
      ]
      \addplot[draw=none, fill=blue!50]
      coordinates {
          (0, 51.7)
          (1, 57.1)
          (2, 60.7)
          (3, 63.7)
          (4, 64.5)
          (5, 65.1)
        };
    \end{axis}
  \end{tikzpicture}
  \caption{The contribution of each part to HOTA metric on DanceTrack validation set.}
  \label{fig:overview_hota}
\end{figure}

\section{Experiments}

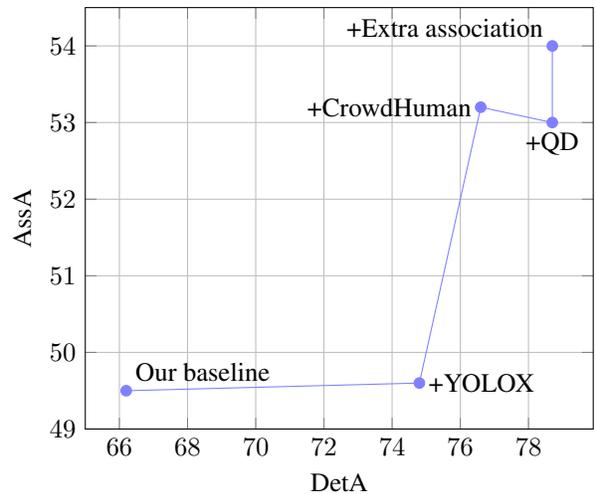
\begin{figure}
  \begin{tikzpicture}
    \begin{axis}[
        width=\linewidth,
        xlabel={DetA},
        ylabel={AssA},
        xmin=65, xmax=79.9,
        ymin=49, ymax=54.5,
        xlabel near ticks,
        ylabel near ticks,
        grid=major,
        enlargelimits=false
      ]

      \addplot[only marks, mark=*, blue!50] table {
          x y
          66.2 49.5
          74.8 49.6
          76.6 53.2
          78.7 53.0
          78.7 54.0
        };
      \draw[draw=blue!50]
      (axis cs:66.2, 49.5) node[anchor=south west]{Our baseline} --
      (axis cs:74.8, 49.6) node[anchor=west]{+YOLOX} --
      (axis cs:76.6, 53.2) node[anchor=east]{+CrowdHuman} --
      (axis cs:78.7, 53.0) node[anchor=north]{+QD} --
      (axis cs:78.7, 54.0) node[anchor=south east]{+Extra association};

    \end{axis}
  \end{tikzpicture}
  \caption{The contribution of each component to AssA and DetA metrics on DanceTrack validation set.}
  \label{fig:overview_assa_deta}
\end{figure}

We evaluate the aforementioned parts in our method on the DanceTrack validation set in this section. The improvements on HOTA, AssA and DetA in each part are summarized in \autoref{fig:overview_hota} and \autoref{fig:overview_assa_deta}.

\subsection{Implementation Details}

For both training and inference, we generate object bounding boxes with YOLOX for each frame individually. For the DanceTrack dataset, we use the pretrained YOLOX model available in the DanceTrack dataset GitHub repo\footnote{https://github.com/DanceTrack/DanceTrack}. The model weight is also used in OC-SORT \cite{cao2022observation}. For the CrowdHuman dataset, we use the YOLOX model for testing on MOT17 provided by ByteTrack \cite{zhang2021bytetrack}.

For MOTR tracker, our baseline implementation is based on the official repo\footnote{https://github.com/megvii-research/MOTR} with several modifications:
\begin{itemize}
  \itemsep0pt
  \item Object bounding boxes are clipped by the frame boundary after data augmentation.
  \item The HSV augmentation applied in the training of the YOLOX detector is adopted.
  \item Both detection and track queries are formulated as anchors. The learnable query PE is replaced with the sine-cosine PE of 4D learnable anchors.
  \item Track queries are filtered by a score threshold of 0.5 during training, which creates false positive (FP) and false negative (FN) track queries to enhance the handling of FPs and FNs during inference. In this way, we do not manually insert negative or drop positive track queries to create hand-crafted FP and FN, \ie, $p_{drop}=0$ and $p_{insert}=0$.
\end{itemize}
These modifications boost HOTA performance by $5.4\%$, as shown in \autoref{tab:impl}.

\begin{table}
  \centering
  \caption{The performance comparison of MOTR official implementation and our baseline on the DanceTrack validation set.}
  \label{tab:impl}
  \begin{tabular}{lccc}
    \toprule
    MOTR          & HOTA & DetA & AssA \\ \midrule
    Official repo & 51.7 & 69.0 & 39.0 \\
    Our baseline  & 57.1 & 66.2 & 49.5 \\ \bottomrule
  \end{tabular}
\end{table}

We train for 5 epochs with a fixed clip size of 5. The training set of the DanceTrack \cite{peize2021dance} dataset and the training and validation set of the CrowdHuman \cite{shao2018crowdhuman} are used for training. The initial learning rate $2\times 10^{-4}$ is dropped by a factor of 10 at the 4$\rm ^{th}$ epoch. If YOLOX is not used, we use 300 learnable anchors with corresponding learnable query embedding as the detect query for MOTR. Otherwise, we use 10 learnable anchors and concatenate them with YOLOX proposals to recall objects missed by the YOLOX detector. For inference, the score threshold for detection is set to 0.5 and the maximum age for negative track queries is set to 20. Other hyperparameters and supervision are the same as the original MOTR.

\subsection{Ablation study}

\paragraph{CrowdHuman joint training and YOLOX proposal}

As shown in \autoref{tab:yolox}, utilizing YOLOX prediction as MOTR detect query initialization \emph{consistently improves all three metrics} (HOTA, DetA, and AssA) regardless of whether the CrowdHuman dataset is used.

It is worth noticing that with the joint training on the CrowdHuman dataset, using YOLOX for proposal generation significantly improves association performance, as indicated by the nearly $10\%$ improvement in AssA.

While using the CrowdHuman dataset alone does not help improve tracking due to the significant drop in association performance, it further improves both the detection and association with the help of YOLOX.

\begin{table}
  \centering
  \caption{Ablation study of CrowdHuman joint training and YOLOX proposal on the DanceTrack validation set.}
  \label{tab:yolox}
  \begin{tabular}{ccccc}
    \toprule
    CrowdHuman   & YOLOX        & HOTA     & DetA     & AssA     \\ \midrule
                 &              & 57.1     & 66.2     & 49.5     \\
                 & $\checkmark$ & 60.7     & 74.8     & 49.6     \\
    $\checkmark$ &              & 56.7     & 73.7     & 43.9     \\
    $\checkmark$ & $\checkmark$ & \bf 63.7 & \bf 76.6 & \bf 53.2 \\ \bottomrule
  \end{tabular}
\end{table}

\paragraph{Query denoising (QD)}

We add query denoising as an auxiliary task to our method. Experimental results in \autoref{tab:qd} show that using query denoising greatly improves detection performance and further improves HOTA by $0.8\%$. Using query denoising with the default scale of noise (0.4) hurts association performance. We attribute this to the gap between denoising and tracking tasks, as the artificial noise is much larger than the cross-frame motion of the instances. Our choice of noise range achieves a 2.1\% improvement in DetA, while almost retaining AssA performance.

\begin{table}
  \centering
  \caption{Ablation study on query denoising on the DanceTrack validation set. The scale of noise $\lambda_1$ and $\lambda_2$ follows the definition in DN-DETR \cite{li2022dn}.}
  \label{tab:qd}
  \begin{tabular}{ccccc}
    \toprule
    $\lambda_1$ & $\lambda_2$ & HOTA & DetA & AssA \\ \midrule
    \multicolumn{2}{c}{No QD} & 63.7 & 76.6 & \bf 53.2 \\
    0.4 & 0.4  & 63.1 & 77.7 & 51.5 \\
    0.1 & 0.05  & \bf 64.5 & \bf 78.7 & 53.0 \\ \bottomrule
  \end{tabular}
\end{table}

\paragraph{Extra Association in post-processing}

We find that keeping tracked queries with negative predictions (lost or occluded) for 20 consecutive frames produces the best HOTA result.
To associate objects that are occluded for more than 20 frames, we design an extra post-processing step that connects the tracks that exit and reappear within 20 to 100 frames so long as the linkage is unique (\ie, there are no entrance or exit of other objects during that time interval).

For the prediction with validation HOTA 64.5 in the previous subsection, the post-processing step connects a total of 21 tracks in the validation set. This operation links the tracks of objects that have been obscured for more than 1 second without affecting the detection performance. The final AssA is improved by $1.0\%$ and HOTA is improved by $0.6\%$ as a result.

\begin{table}
  \centering
  \caption{Ablation study on extra association in post-processing on the DanceTrack validation set.}
  \label{tab:post}
  \begin{tabular}{cccc}
    \toprule
    Extra Association & HOTA & DetA & AssA \\ \midrule
                      & 64.5 & 78.7 & 53.0 \\
    $\checkmark$      & 65.1 & 78.7 & 54.0 \\ \bottomrule
  \end{tabular}
\end{table}

\section{Final Result}

\begin{table}
  \centering
  \caption{The final leaderboard of the group dance challenge. We ranked 1$\rm ^{st}$ on the DanceTrack test set. The individual rankings for each metric are listed in parentheses.}
  \label{tab:final}
  \begin{tabular}{llccc}
    \toprule
    Ranking & User    & HOTA         & DetA      & AssA         \\ \midrule
    \bf 1   & \bf mfv & \bf 73.4 (1) & 83.7 (2)  & \bf 64.4 (1) \\
    2       & C-BIoU  & 66.6 (2)     & 81.3 (13) & 54.7 (2)     \\
    3       & ymzis69 & 64.6 (3)     & 82.5 (3)  & 50.7 (3)     \\ \bottomrule
  \end{tabular}
\end{table}

The final result of the group dance challenge is shown in \autoref{tab:final}.
Our learning-based solution shows a much stronger association performance (+9.7\% AssA and +6.8\% HOTA) compared to the matching-based solutions applied by C-BIOU and ymzis69. Note that the model for submission is jointly trained on the training and validation sets of DanceTrack (+1\% HOTA), and four models are ensembled together (+ 2\% HOTA).

  {\small
    \bibliographystyle{ieee_fullname}
    \bibliography{egbib}
  }

\end{document}